\documentclass[11pt]{article}
\usepackage[utf8]{inputenc}
\usepackage[]{geometry}

\usepackage{authblk}
\usepackage{graphicx}
\usepackage{caption}
\usepackage{subcaption}
\usepackage[english]{babel}
\usepackage{csquotes}
\usepackage{amsmath}
\usepackage{amssymb}
\usepackage{bm}
\usepackage[section]{placeins}
\usepackage{nameref}

\usepackage[hyphenbreaks]{breakurl}
\usepackage[hyphens]{url}
\usepackage[hidelinks]{hyperref}
\usepackage{natbib}
\usepackage[ruled]{algorithm2e}
\usepackage{mathpazo}
\usepackage{xurl}

\usepackage{orcidlink}

\setcitestyle{authoryear, open={(},close={)}}

\providecommand{\keywords}[1]
{
  \small	
  \textbf{Keywords:} #1
}

\title{On the Creativity of Large Language Models}
\author[1]{Giorgio Franceschelli\,\orcidlink{0000-0003-3210-3015}}
\author[2, 1]{Mirco Musolesi\,\orcidlink{0000-0001-9712-4090}}
\affil[1]{University of Bologna, Italy}
\affil[2]{University College London, United Kingdom}
\affil[ ]{giorgio.franceschelli@unibo.it, m.musolesi@ucl.ac.uk}
\date{}

\begin{document}

\maketitle

\begin{abstract}
   Large Language Models (LLMs) are revolutionizing several areas of Artificial Intelligence. One of the most remarkable applications is creative writing, e.g., poetry or storytelling: the generated outputs are often of astonishing quality. However, a natural question arises: can LLMs be really considered creative? In this article, we first analyze the development of LLMs under the lens of creativity theories, investigating the key open questions and challenges. In particular, we focus our discussion on the dimensions of value, novelty, and surprise as proposed by Margaret Boden in her work. Then, we consider different classic perspectives, namely product, process, press, and person. We discuss a set of ``easy'' and ``hard'' problems in machine creativity, presenting them in relation to LLMs. Finally, we examine the societal impact of these technologies with a particular focus on the creative industries, analyzing the opportunities offered, the challenges arising from them, and the potential associated risks, from both legal and ethical points of view.
\end{abstract}


%\subsection*{Acknowledgement}

%Acknowledgments go here.

%\subsection*{Conflict of Interest}

%The authors declare that there is no conflict.

\section{Introduction}\label{sec1}

Language plays a vital role in how we think, communicate, and interact with others\footnote{As remarked by ChatGPT itself when asked about the importance of language.}. It is therefore of no surprise that natural language generation has always been one of the prominent branches of artificial intelligence \citep{jurasky23}. We have witnessed a very fast acceleration of the pace of development in the past decade culminated with the invention of transformers \citep{vaswani17}. The possibility of exploiting large-scale data sets and the availability of increasing computing capacity has led to the definition of the so-called foundation models, which are able to achieve state-of-the-art performance in a variety of tasks \citep{bommasani21}. 

Among them, large language models (LLMs) are indeed one of the most interesting developments. They have captivated the imagination of millions of people, also thanks to a series of entertaining demonstrations and open tools released to the public. The examples are many from journal articles\footnote{\url{www.theguardian.com/commentisfree/2020/sep/08/robot-wrote-this-article-gpt-3}} to culinary recipes \citep{lee20} and university-level essays\footnote{\url{https://www.theguardian.com/technology/2022/dec/04/ai-bot-chatgpt-stuns-academics-with-essay-writing-skills-and-usability}}. LLMs have also been used to write papers about themselves writing papers \citep{gpt22}.
%They have also been used
They are commonly used for creative tasks like poetry or storytelling and the results are often remarkable\footnote{See, for instance: \url{https://www.gwern.net/GPT-3}}. Notwithstanding, it is not obvious whether these ``machines'' are truly creative, at least in the sense originally discussed by Ada Lovelace \citep{menabrea43}. LLMs have already been analyzed (and sometimes criticized) from different perspectives, e.g., fairness \citep{bender21}, concept understanding \citep{bender20}, societal impact \citep{tamkin21}, and anthropomorphism \citep{shanahan2024talking} just to name a few. However, a critical question has not been considered yet: \textit{can LLMs be considered creative}?

By taking into account classic frameworks for analyzing creativity, such as Boden's three criteria \citep{boden03} and other prominent cognitive science and philosophical theories (e.g., \cite{amabile83,csikszentmihalyi88,gaut10}), we will try to answer this question. We will discuss the dimensions according to which we believe LLMs should be analyzed in order to evaluate their level of machine creativity. To the best of our knowledge, this article represents one of the first investigations of the problem of LLM creativity from a theoretical and philosophical perspective.

The remainder of the paper is structured as follows. First, we briefly review the past developments in automatic text generation and artificial creativity (Section \ref{textgeneration}) that led to today's LLMs. Then, we analyze LLMs from the perspective of Boden's three criteria (Section \ref{bodencriteria}), as well as considering other relevant philosophical theories (Section \ref{fourp}). Finally, we discuss the practical implications of LLMs for the arts, creative industries, design, and, more in general, scientific and philosophical inquiry (Section \ref{implications}). Section \ref{conclusion} concludes the paper, outlining the open challenges and a research agenda for future years.

\section{A Creative Journey from Ada Lovelace to Foundation Models} \label{textgeneration}

It was the year 1843 when Ada Lovelace wrote that the Analytical Engine \citep{babbage64} \textit{``has no pretensions to originate anything. It can do whatever we know how to order it to perform''} \citep{menabrea43}. This statement was then defined as ``Lovelace's objection'' by Alan Turing, who also provided an alternative formulation: a machine can never ``take us by \textit{surprise}'' \citep{turing50}.
This was just the beginning of an ongoing philosophical discussion, which has often included psychological elements, around human creativity \citep{barron55,berlyne60,bruner62,newell62,stein74}, as well as computational creativity \citep{macedo04,wiggins06,jordanous09,boden09,maher10,colton2012computational}.

In general, computer scientists have always been fascinated by the possibility of building machines able to express themselves through writing, e.g., by composing poems and short stories, creating paintings, and so on. In particular, the rise of automatic text generation was contextual to the birth of personal computers. Examples include the Computerized Haiku by Margaret Masterman\footnote{\url{http://www.in-vacua.com/cgi-bin/haiku.pl}}, the storyteller TALE-SPIN \citep{meehan77}, Racter and its poems' book \citep{racter84}, and UNIVERSE, which was able to generate coherent and consistent characters \citep{lebowitz83}, just to name a few. Different techniques have been explored, from planning (e.g., \cite{riedl10}) and case-based reasoning (e.g., \cite{turner94}) to evolutionary strategies (e.g., \cite{manurung12}). Some approaches combine all of them together \citep{gervas13}. 

Only with the advent of neural networks and learning systems, we observed a real step-change.
In particular, deep language models, i.e., probabilistic models of in-context token occurrences trained on a corpus of text with deep learning, easily allow the sampling of new text, facilitating and automating natural language generation. For instance, recurrent neural networks with long-short term memory (LSTM) \citep{hochreiter97} or gated-recurrent units (GRUs) \citep{cho14} can predict next character \citep{karpathy15}, word \citep{potash15}, syllable \citep{zugarini19}, or event \citep{martin18} given previous ones, allowing to compose text that spans from short movie scripts to knock-knock jokes \citep{miller19}. Other successful generative methods include generative adversarial networks (GANs) \citep{yu16,zhang17} and variational auto-encoders (VAEs) \citep{bowman16,semeniuta17}. We refer the interested reader to \cite{franceschelli2024creativity} for an in-depth survey of deep learning techniques applied to creative artifacts.

These models tend to scale poorly to long sequences, and they are often unable to capture the entire context. For this reason, current state-of-the-art language models make use of attention \citep{bahdanau15} and transformers \citep{vaswani17}. In recent years, several models based on these mechanisms have been proposed. They usually rely on a very large number of parameters and are trained on corpus datasets of greater and greater size \citep{devlin18,radford2019language,shoeybi19,brown20,raffel20,rosset20,rae21,chowdhery2023palm,du22,hoffmann22,smith22,thoppilan22}.
Thanks to in-context learning techniques such as zero-shot or few-shot learning \citep{dong2024survey}, these models can produce more specific and specialized content, such as poems or stories \citep{swanson21}, by simply providing a description of the task and possibly some examples. However, finding the correct input and high-quality demonstrations for solving this type of task can be challenging \citep{liu2022makes}. Certain domains might require more fine-grained knowledge than that acquired during pre-training \citep{peng2023does}. Because of this, other methods to adapt a pre-trained model have been considered.
%Although such models can be used with zero-shot or few-shot learning to produce poems and stories \citep{swanson21}\gf{COMMENT 2 - Discuss more}, the problem of adaptation has been central to their development. 
LLMs can involve re-training through plug-and-play attribute classifiers \citep{dathathri20}; re-training to produce paragraphs coherent with a given outline \citep{rashkin20}; fine-tuning with specific corpora for writing specific text \citep{sawicki22,wertz22}; or fine-tuning to maximize human preferences \citep{ziegler19} or to generate specific literary outputs, such as poetry \citep{pardinas23}. 
Nevertheless, the recent advancements in LLMs can be attributed to the introduction of fine-tuning through reinforcement learning from human feedback (RLHF) \citep{christiano2017deep}. It consists of three steps: fine-tuning the pre-trained model in a supervised fashion on human-produced answers to sampled questions; training a reward model to predict which text among different options is the most appropriate based on human-labeled rankings; and fine-tuning the language model to maximize the learned reward \citep{stiennon20}. Although the main goal of RLHF is to improve conversational skills while mitigating mistakes and biases, it has also led to models capable of producing on-demand poems, songs, and novels, gaining global popularity\footnote{\url{https://www.forbes.com/sites/martineparis/2023/02/03/chatgpt-hits-100-million-microsoft-unleashes-ai-bots-and-catgpt-goes-viral/?sh=70994247564e}}. Based on RLHF, first ChatGPT\footnote{\url{https://openai.com/blog/chatgpt/}} and then GPT-4 paved the way for several other similar models: Google’s Gemini \citep{gemini2023gemini}, which extends to multimodal data; Meta’s Llama models \citep{dubey2024llama,touvron2023llama}, which replace RLHF with the more efficient direct preference optimization (DPO) \citep{rafailov2023direct}; Mixtral \citep{jiang2024mixtral}, which adaptively selects its layers’ parameters from distinct groups to increase the total parameter count without raising computational costs; and many others, as the competition intensifies day by day \citep{zhao2023survey}. While they may differ in some technical details, these LLMs are always pre-trained on vast, general corpora of data and then fine-tuned using some form of RLHF to enhance their conversational skills.

\section{Large Language Models and Boden's Three Criteria} \label{bodencriteria}

Margaret Boden defines creativity as ``the ability to come up with ideas or artifacts that are \textit{new}, \textit{surprising} and \textit{valuable}'' \citep{boden03}. In other words, Boden implicitly derives criteria that can be used to identify a creative \textit{product}. They suggest that creativity is about \textit{novelty}, \textit{surprise} and \textit{value}.
We will refer to them as Boden's three criteria. 
In the following, we will analyze to what extent state-of-the-art LLMs satisfy them and we will question if LLMs can be really considered creative.

Value refers to utility, performance, and attractiveness \citep{maher10}. It is also related to both the quality of the output, and its acceptance by society. Due to the large impact LLMs are already having \citep{bommasani21} and the quality of outputs of the systems based on them \citep{stevenson22}, it is possible to argue that the artifacts produced by them are indeed valuable.

Novelty refers to the dissimilarity between the produced artifact and other examples in its class \citep{ritchie07}. However, it can also be seen as the property of not being in existence before. This is considered in reference to either the person who came up with it or the entire human history. The former is referred to as psychological creativity (shortened as \textit{P-creativity}), whereas the latter is historical creativity (shortened as \textit{H-creativity}) \citep{boden03}. While the difference appears negligible, it is substantial when discussing LLMs in general. Considering these definitions, a model writing a text that is not in its training set would be considered as P-novel, but possibly also H-novel, since LLMs are commonly trained on all available data. Their stochastic nature and the variety of prompts that are usually provided commonly lead to novel outcomes \citep{mccoy2023much}; LLMs may therefore be capable of generating artifacts that are also new. However, one should remember how such models learn and generate. LLMs still play a sort of \textit{imitation game}, without a focus on (computational) novelty \citep{fazi19}. Even if prompted with the sentence ``I wrote a new poem this morning:'', they would nonetheless complete it with what is most likely to follow such words, e.g., something close to what others have written in the past \citep{shanahan2024talking}. It is a probabilistic process after all. The degree of dissimilarity would therefore be small \textit{by design}. High values of novelty would be caused either by accidental, out-of-distribution productions or by careful prompting, i.e., one that would place the LLM in a completely unusual or unexpected (i.e., novel) situation.

Surprise instead refers to how much a stimulus disagrees with expectation \citep{berlyne71}. It is possible to identify three kinds of surprise, which correspond to three different forms of creativity. \textit{Combinatorial creativity} involves making unfamiliar combinations of familiar ideas. \textit{Exploratory creativity} requires finding new, unexplored solutions inside the current style of thinking. \textit{Transformational creativity} is related to changing the current style of thinking \citep{boden03}. These three different forms of creativity involve surprise at increasing levels of abstraction: combining existing elements, exploring for new elements coherent with the current state of the field, and transforming the state of the field to introduce other elements. The autoregressive nature of classic LLMs makes them unlikely to generate surprising products \citep{bunescu19} since they are essentially trained to follow the current data distribution \citep{shanahan2024talking}. By relying only on given distributions and being trained on them, LLMs might at most express combinatorial or exploratory creativity. Of course, specific different solutions may be generated by means of prompting or conditioning. For instance, recent LLMs can write poems about mathematical theories, a skill that requires the application of a certain existing style to a given topic, yet leading to new and unexplored solutions. However, the result would hardly be unexpected for whom has prompted the text. For an external reader, the surprise would probably come from the idea of mathematical theories in verses, which is due to the user (or by the initial astonishment of a machine capable of it \citep{waite19}). Transformational creativity is not achievable through the current LLM training solutions. In theory, other forms of training or fine-tuning might circumvent this limitation, allowing the model to forget the learned rules in order to forge others. However, this is not the case with current models.
ChatGPT and all the other state-of-the-art LLMs introduced in Section \ref{textgeneration} are fine-tuned with RLHF or DPO.
While in theory this could lead to potentially surprising generation, its strict alignment to very careful and pre-designed human responses leads to the generation of text that tends to be less diverse \citep{kirk2024understanding} and that might be considered \textit{banal} \citep{hoel22}.

Nonetheless, the outputs from such models are often considered creative by the person interacting with them or exposed to their best productions. Though this is apparently in contrast with what was discussed above, we can explain this phenomenon by considering the fact that our perception does not usually align with theoretical definitions of creativity. Indeed, we do not typically judge the creativity of a product by considering its potential novelty and surprise in relation to its producer, but rather in relation to ourselves. Something can be new for the beholder, leading to a new kind of novelty which we call \textit{B-novelty}, as it is the one ``in the eye of the beholder'', but not new for the producer nor the entire human history. The same applies to surprise: a product can violate the observer's expectations in many ways without being unexpected considering the entire domain. In other words, the product of an LLM can appear to be creative - or be B-creative - even if it is not \textit{truly} creative according to the theory of creativity.

In conclusion, while LLMs are capable of producing artifacts that are valuable, achieving P- or H-novelty and surprise appears to be more challenging. It is possible to argue that LLMs may be deemed able to generate creative products if we assume the definition of combinatorial creativity. To achieve transformational creativity, alternative learning architectures are probably necessary; in fact, current probabilistic solutions are intrinsically limiting in terms of expressivity. We believe that this is a fundamental research area for the community for the years to come.

\section{Easy and Hard Problems in Machine Creativity}\label{fourp}

LLMs might be able to generate creative products in the future. However, the fact that they will be able to generate these outputs will not make them intrinsically creative.
Indeed, as \cite{floridi20} puts it, it is not \textit{what} is achieved but \textit{how} it is achieved that matters. An interesting definition that considers both the \textit{what} and \textit{how} dimensions is the one from \cite{gaut03}: creativity is the capacity to produce original and valuable items by \textit{flair}. Exhibiting flair means exhibiting a relevant purpose, understanding, judgment, and evaluative abilities. Such properties are highly correlated with those linked with \textit{process}, i.e., motivation, perception, learning, thinking, and communication \citep{rhodes61}. Motivation is a crucial part of creativity, as it is the first stage of the process. Usually, it comes from an intrinsic interest in the task, i.e., the activity is interesting and enjoyable for its own sake \citep{deci85}. However, LLMs lack the intention to write. They can only deal with ``presented'' problems, which are less conducive to creativity \citep{amabile96}. The process continues with the preparation step (reactivating store of relevant information and response algorithms), the response generation, and its validation and communication \citep{amabile83}. The last two steps allow one to produce different response possibilities and to internally test them in order to select the most appropriate. Again, LLMs do not contain such a self-feedback loop. At the same time, they are not trained to directly maximize value, novelty, or surprise. They only output content that is likely to follow given a stimulus in input \citep{shanahan2024talking}. In other words, they stop at the first stage of creative learning, i.e., imitation, not implementing the remaining ones, i.e., exploration and intentional deviation from conventions \citep{riedl18}.

However, paraphrasing Chalmers \citep{chalmers1996}, these appear as \textit{easy} problems to solve in order to achieve creativity, since solutions to them can be identified by taking into consideration the underlying training and inference processes. The \textit{hard} problem in machine creativity is about the intentionality and the self-awareness of the creative process in itself. Even though the intent of running the LLM may be achieved by its outcome, it is in an unintentional way \citep{terzidis22}; as current generative AI models are only causal, and not intentional, agents \citep{johnson19}.
Indeed, a crucial aspect of the creative process is the perception and the ability of \textit{self-evaluating} the generated outputs \citep{amabile83}. This can be seen as a form of creative self-awareness. While not strictly necessary to generate a response, this ability is essential in order to self-assess its quality, so as to correct it or to learn from it. However, no current LLM is able to self-evaluate its own responses. 
LLMs can in theory recognize certain limitations of their own texts after generating them, e.g., by ranking them \citep{franceschelli2024creative} or by assigning quality- and diversity-based scores \citep{bradley2024quality}. Then, they can try to correct, modify, or rephrase the outputs if asked to do so (i.e., through an external intervention). However, they would do it only by guessing what is the most likely re-casting of such responses or through the application of a set of given rules. It is worth noting that this is something distinct from the problem of the potential emergence of theory of mind in these systems \citep{bubeck2023sparks}.

Indeed, product and process are not sufficient to explain creativity. \cite{rhodes61} theorizes that four perspectives have to be considered: product (see Section \ref{bodencriteria}) and process (discussed above), but also the so-called \textit{press} and \textit{person}.
Press refers to the relationship between the product and the influence its environment has upon it \citep{rhodes61}. Individuals and their works cannot be isolated from the social and historical milieu in which their actions are carried out. Products have to be accepted as creative by the society, and producers are influenced by the previously accepted works, i.e., the domain \citep{csikszentmihalyi88}. The resulting system model of creativity is a never-ending cycle where individuals always base their works on knowledge from a domain, which constantly changes thanks to new and valuable artifacts (from different individuals). For example, individuals generate new works based on the current domain; the field (i.e., critics, other artists, the public, etc.) decides which of those works are worth promoting and preserving; the domain is expanded and, possibly, transformed by these selected works; individuals generate new works based on the updated current domain; and then this cycle repeats. 

However, LLMs cannot currently adapt through multiple iterations in the way described above; they just rely on one, fixed version of the domain and generate works based on it. The current generation of LLMs are \textit{immutable}  entities, i.e., once the training is finished, they remain frozen reflecting a specific state of the domain. In other words, they are not able to adapt to new changes.
In-context learning can simulate an adaptation to new states of the domain. The constantly increasing context length \citep{hsieh2024ruler} allows researchers to provide more and more information to LLMs without re-training them, although a longer context might lead to performance degradation \citep{li2024longcontext}. This enables the representation of the current state of the domain through an adequate prompt, allowing the model to generate different outputs according to environmental changes. For example, in \cite{park2023generative}, multiple LLM-based agents interact through natural language in a sandbox environment inspired by \textit{The Sims}. Each agent stores, synthesizes, and applies relevant memories to generate believable behavior through in-context learning, leading to emergent social behaviors. The study of emergent behaviors of LLM-based agents at the population level is an active research area \citep{guo2024large}. It is easy to imagine the simulation of creative or artistic environments, such as a virtual multi-agent translation company \citep{wu2024perhaps}, as well.

However, LLMs are like the main character of \textit{Memento}: they always possess all the capabilities, but each time they ``wake up'', they need to re-collect all the information about themselves and their world. The time - or space - to acquire such information is limited, and by the next day, they will have forgotten it all. In other words, these generative agents do not truly adapt or learn new things about the changing domain. Placing them in a different environment that requires a different prompt will make them start over, without the possibility of leveraging previously acquired experience.

On the other hand, fine-tuning actually updates network weights, but it requires a potentially large training dataset. Indeed, several current research efforts are in the direction of introducing adaptation for specific domains, tasks, cultural frameworks, and so on.
In order to be able to be part of the never-ending creative cycle mentioned above, LLMs should constantly adapt. Continual learning \citep{kirkpatrick17,shin17} for LLMs \citep{sun20,wu22} represents a promising direction, yet unexplored for creative applications.

Finally, person covers information about personality, intellect, temperament, habits, attitude, value systems, and defense mechanisms \citep{rhodes61}. While several of the properties of press and process might be achieved - or at least simulated - by generative learning solutions, those related to the creative person appear out of discussion \citep{browning23}.
Several works have analyzed whether LLMs can pass tests intended to evaluate human psychological skills \citep{binz2023using,macmillan2024ir,stevenson2022putting}, sometimes with promising results \citep{kosinski2024evaluating,lampinen2024language}. However, according to the best-supported neuroscientific theories of consciousness, current AI systems are not conscious \citep{butlin2023consciousness}.
As \cite{ressler23} pointed out, LLMs have no self to which to be true when generating text and are intrinsically unable to behave authentically as individuals. They merely ``play the role'' of a character or, more accurately, a superposition of simulacra within a multiverse of possible characters induced by their training \citep{shanahan2023role,shanahan2024simulacra}. This results in a perceived self-awareness, stemming from our inclination to anthropomorphize \citep{deshpande2023anthropomorphization,seth2021being}.
In conclusion, all the properties listed above require some forms of consciousness and self-awareness, which are difficult to define in themselves and are related to the \textit{hard} problem introduced before.
Creative-person qualities in generative AI might eventually be the ultimate step in achieving human-like intelligence.

\section{Practical Implications} \label{implications}

The application of large language models to fields like literature or journalism opens up a series of practical questions. Since LLMs can be used to produce artifacts that would be protected if made by humans, a first concern is the definition of legal frameworks in which they will be used. Copyright for generative AI is currently a hotly debated topic \citep{guadamuz2017androids,franceschelli22,lee2024talkin,miernicki21}, due to the fact that current laws do not contemplate works produced by non-human beings (with few notable exceptions \citep{bond19}).
Copyright applies to creative \textit{works of authorship} (as referred to in the US Copyright Code), i.e., works showing a minimum degree of originality \citep{gervais02} and reflecting author's personality \citep{deltorn17}. As discussed earlier, current LLMs might satisfy the first condition, but they cannot be considered creative persons, therefore missing the latter requirement. For this reason, works produced by LLMs can be protected if and only if the original contribution is provided by a human, e.g., the user who writes the prompt that is used as input of the model, who in turn will be the rights holder.
The definition of the criteria for classifying a source of originality is a fundamental problem since there is a clear need to discriminate between protected and publicly available works.

While a higher degree of novelty is unnecessary for claiming protection, it might be crucial for other legal aspects. In particular, LLMs are trained in a supervised fashion on real data, which also include protected works \citep{bandy2021addressing}. Apart from questions upon the legitimacy of such training \citep{franceschelli22}, LLMs may learn to reproduce portions of them \citep{liang22} because of the memorization of training data \citep{carlini2023quantifying}. This would violate their reproduction or adaptation right \citep{bonadio20}. A different, creative-oriented training approach should mitigate such risk, also facilitating fair-use doctrine application \citep{asay20}.

Whether or not LLM works obtain protection, we believe their societal impact will be tremendous (see also \cite{newton23}). We have a positive view in terms of the applications of LLMs, but there are intrinsic risks related to their adoption. It is apparent that since LLMs are able to write articles or short stories, as the quality of their inputs gets better and better, there is the risk that certain jobs in the professional writing industry will essentially disappear \citep{poncedelcastillo23,tamkin21}.
However, we must remind that current LLMs are not as reliable as humans, e.g., they cannot verify their information and they can propagate biases from training data. In addition, the quality of the output strictly depends on the prompt, which might in turn demand human skills and more time. Writers can be threatened as well. Though not in violation of copyright, LLMs may exploit certain ideas from human authors, capitalizing on their efforts in ways that are less expensive or time-consuming \citep{weidinger22}. The questionable creative nature of LLMs discussed so far might suggest artificial works to be of less quality than humans, therefore not providing a real threat. Nonetheless, more creative LLMs would diverge more consistently from existing works, reducing the risk of capitalizing on others' ideas. The lack of current copyright protection for generated works can also foster such replacements for tasks where a free-of-charge text would be preferable to a high-quality (but still costly) one. Finally, one last threat may be posed by human and artificial works being indistinguishable \citep{dehouche21}. The users obtaining such outputs might therefore claim them as the authors, e.g., for deceiving readers \citep{grinbaum22}, for cheating during exams \citep{fyfe23}, or for improving bibliometric indicators \citep{crothers22}. Mitigation of such threats through dedicated policies\footnote{\url{https://bigscience.huggingface.co/blog/the-bigscience-rail-license}} or designed mechanisms of watermarks \citep{kirchenbauer2023watermark} are already being developed.

However, as we said, we believe that, overall, the impact of these technologies will be positive. LLMs also provide several opportunities for creative activities.
Given their characteristics, humans are still required, especially for prompting, curation, and pre-/post-production. This means that the role of writers and journalists may be transformed, but not replaced. On the contrary, LLMs provide new opportunities for humans, who will be able to spend more time validating news or thinking up and testing ideas. LLMs can also adapt the same text to different styles (see combinatorial creativity in Section \ref{bodencriteria}): by doing so, an artifact can be adapted to reach wider audiences. In the same way, LLMs also represent a valuable tool in scientific research \citep{fecher23}, especially for hypothesis generation \citep{gero22}.

Indeed, we believe that LLMs can also foster human-AI co-creativity \citep{lee22}, since they can be used to write portions of stories in order to serve specific purposes, e.g., they can typify all the dialogues from a character, or they can provide more detailed descriptions of scenes \citep{calderwood20}. Dialogue systems based on LLMs can be used for brainstorming. In the same way, the generated responses may augment writers' inherently multiversal imagination \citep{reynolds21}. LLMs can also represent a source of inspiration for plot twists, metaphors \citep{chakrabarty23}, or even entire story plans \citep{mirowski22}, even though they sometimes appear to fail in accomplishing these tasks at human-like level \citep{ippolito22}. Being intrinsically powerful tools, through human-AI co-creation, LLMs may eventually allow the development of entire new arts, as has been the case for any impactful technology in the past centuries \citep{eisenstein79,silva22}.

\section{Conclusion} \label{conclusion}

The latest generation of LLMs is attracting increasing interest from both AI researchers and the general public due to the astonishing quality of their productions. Questions naturally arise around the actual creativity of these technologies.
In this paper, we have discussed whether or not LLMs can actually be deemed as creative; we started by considering Boden's three criteria, i.e., value, novelty, and surprise. While LLMs are capable of value and a weak version of novelty and surprise, their inner autoregressive nature seems to prevent them from reaching transformational creativity. Then, we have examined perspectives beyond the creativity of their products. A creative process would require motivation, thinking, and perception, properties that current LLMs do not possess. The social dimension of creativity (usually referred to as the press) would demand to be placed in and influenced by a society of creative agents, requiring LLMs adaptive abilities that are only at a very initial stage. We have also framed the problem of creativity in LLMs, and, more in general, machine creativity, in terms of easy problems, i.e., the technical advancements that will be needed to support the algorithmic generation of outputs and the intrinsic hard problem of introducing forms of self-awareness in the creation process itself.

In addition, we have also investigated the practical implications of LLMs and their creative role, considering both legal and societal impacts. In fact, the current legal framework does not appear to be completely suited to the fast-moving field of generative AI. Moreover, the impact of these technologies on creative professions and the arts is difficult to forecast at this stage, but will definitely be considerable.
However, LLMs also provide opportunities for writers, especially in terms of human-AI cooperation. Specific fine-tuning techniques might help LLMs diversify productions and explore the conceptual space they learn from data. Continual learning can enable long-term deployments of LLMs in a variety of contexts. While, of course, all these techniques would only simulate certain aspects of creativity, whether this would be sufficient to achieve artificial, i.e., non-human, creativity, is up to the humans themselves. 

\bibliography{biblio.bib}

\end{document}